\documentclass[aps,pre,reprint,amsmath,amssymb,dvipdfmx]{revtex4-2}
\usepackage{txfonts}
\usepackage{algorithmic}
\usepackage{algorithm}
\usepackage{graphicx}
\usepackage{bm}
\usepackage{url}
\usepackage{mathrsfs}
\usepackage{listings}
\usepackage{color}

\begin{document}

\title{Neural network initialization with nonlinear characteristics and information on hierarchical features}

\author{Hikaru Homma and Jun Ohkubo}

\affiliation{Graduate School of Science and Engineering, Saitama University, Sakura, Saitama 338--8570, Japan}

\begin{abstract}
Initialization of neural network parameters, such as weights and biases, has a crucial impact on learning performance; if chosen well, we can even avoid the need for additional training with backpropagation. For example, algorithms based on the ridgelet transform or the SWIM (sampling where it matters) concept have been proposed for initialization. On the other hand, some works show hierarchical features in trained neural networks; neural networks tend to learn coarse information in the early-stage hidden layers. In this work, we investigate the effects of utilizing information on the hierarchical features in the initialization of neural networks. Hence, we propose a framework that adjusts the scale factors in the SWIM algorithm to capture low-frequency components in the early-stage hidden layers and to represent high-frequency components in the late-stage hidden layers. Numerical experiments on a one-dimensional regression task and the MNIST classification task demonstrate that the proposed method outperforms the conventional initialization algorithms. This work clarifies the importance of intrinsic hierarchical features in learning neural networks, and the finding yields an effective parameter initialization strategy that enhances their training performance.

\end{abstract}

\maketitle

\section{Introduction}\label{sec:introduction}

The fields of machine learning and artificial intelligence are deeply connected to physics, as evidenced by the fact that neural networks won the 2024 Nobel Prize. However, its significant computational cost during both training and utilization has also become problematic from an environmental impact perspective. In utilizing trained neural networks, techniques such as pruning and quantization are widely employed to improve efficiency; see the recent reviews in \cite{Liang2021,Cheng2023}. Here, we focus on the training stage of neural networks from the perspective of sampling, which would also be interesting in statistical physics.

It has been widely known that neural networks exhibit high versatility and strong approximation capability \cite{Cybenko1989,Barron1993}. Their nonlinearity yields powerful and practical trained networks in various applications in image recognition, natural language processing, and speech recognition. However, as the number of hidden layers increases, the number of parameters grows exponentially, leading to longer training times \cite{Sun2016,Sze2017,Sankararaman2020,Abdolrasol2021}. Since training involves non-convex optimization, there is also a risk of local minima and stagnation in flat regions. These issues highlight the critical importance of parameter initialization.

Since conventional backpropagation requires huge computational costs, other types of neural networks have been proposed. One example is the extreme learning machine (ELM) \cite{Huang2004,Huang2006}. The ELM is a training algorithm for a feedforward neural network with only one hidden layer. In the ELM, one randomly chooses hidden node parameters, and the learning process is applied only to the output layer through simple linear regression. Hence, the learning process for the ELM is fast and efficient; for details of the ELM, see the recent review paper \cite{Wang2022}. However, even shallow neural networks suffer from unstable learning. 

To address this problem, one approach is to modify the random initialization of parameters. That is, one could employ some theories and data for the parameter initialization. For example, one can transform shallow neural networks into an integral representation and apply the sampling method called oracle sampling \cite{Murata1996, Sonoda2014}. In the oracle sampling algorithm, hidden-layer parameters are sampled from a ridgelet-transform-based distribution, and the conventional linear regression determines output parameters. In \cite{Homma2024}, we revisited the oracle sampling algorithm from the perspective of importance sampling, which leads to an initialization method that demonstrates higher performance than the original one. These approaches achieve near-trained performance without backpropagation in some cases, and the theoretical framework is also used in discussions of deep networks, utilizing, for example, harmonic analysis on groups \cite{Sonoda2019,Sonoda2023, Sonoda2025}. However, the practical sampling method is limited to single hidden-layer models.

The SWIM (sampling where it matters) algorithm \cite{Bolager2023} was proposed in a different context from the ridgelet-transform-based initialization method, which enables the generation of parameters for deep networks. The SWIM algorithm constructs hidden-layer weights and biases directly from data pairs, determining all parameters, except those of the output-layer, via sampling. The output layer is then fitted by linear regression, allowing high performance even in deep networks without gradient-based training. The SWIM algorithm has also been applied to graph neural networks and higher-order partial differential equations, demonstrating its potential as a non-gradient learning method based on random features \cite{Datar2024, Atamert2025}. The concept was also extended to the boosting method to improve the performance of deep feed-forward networks \cite{Zozoulenko2025} and recurrent neural networks \cite{Bolager2025}.

On the other hand, hierarchical features in deep neural networks have been extensively studied, particularly in convolutional neural networks (CNNs). The CNNs are known to learn low-frequency and coarse features in the early-stage hidden layers, while capturing high-frequency and fine-grained features in the late-stage hidden layers. For example, some studies investigated the hierarchical features in trained neural networks on image recognition \cite{Zeiler2014,Springenberg2015}. Note that the characteristic is different from the frequency principle (F-Principle) \cite{Rahaman2019,Luo2019,Xu2019}, in which the dominant low-frequency components are quickly captured in the learning process. Here, we focus on the bias in the hierarchical features of trained networks.

To analyze such hierarchical representations in CNNs, various visualization techniques have been proposed, including decoder and deconvnet-based methods \cite{Qin2018}. These approaches enable direct observation of how features evolve across layers and have provided strong evidence that CNNs progressively encode information from low to high frequencies.

In the present paper, we propose a parameter initialization method that employs the nature of nonlinear characteristics and information on the hierarchical features. In the SWIM algorithm, we find that the nonlinearity used in the activation function is controllable by changing hyperparameters for each hidden layer. It is possible to reflect the information of the hierarchical features in the setting of the hyperparameters. We confirm that the proposed initialization method improves the performance in a simple one-dimensional example and the MNIST datasets.

The remainder of this paper is structured as follows. Section~\ref{sec:background} provides a brief overview of the SWIM algorithm and presents simple numerical experiments concerning the hierarchical features. Section~\ref{sec:proposal} discusses the nonlinearity and hyperparameters in the SWIM algorithm and proposes a parameter initialization method to incorporate information on the hierarchical features. Section~\ref{sec:experiments} presents numerical results demonstrating the effectiveness of the proposed method. Section~\ref{sec:conclusion} yields some considerations and conclusions.

\section{Background knowledge}
\label{sec:background}

\subsection{Sampling Where It Matters (SWIM)}
\label{sec:background_swim}

\begin{figure}[b]
  \centering
  \includegraphics[width=0.48\textwidth]{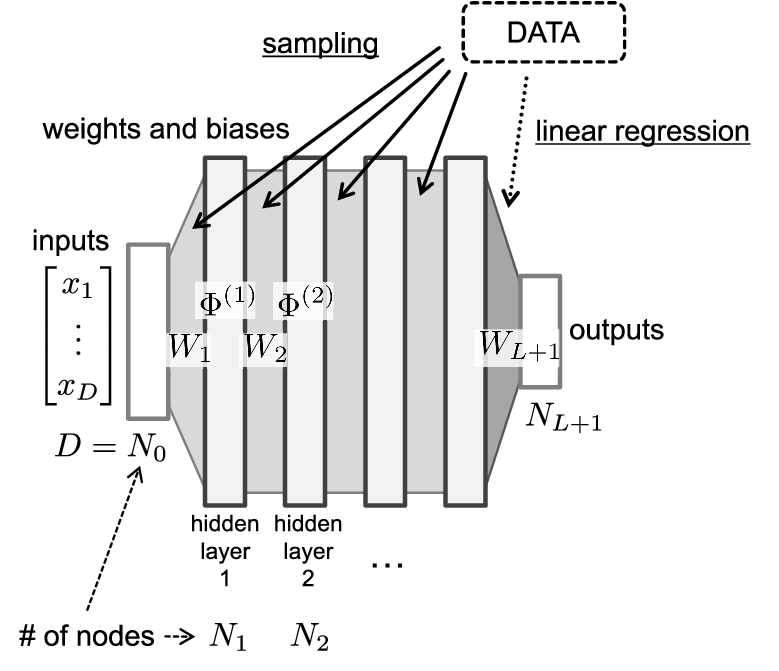}
  \caption{The parameter initialization in the SWIM algorithm. The parameters in the hidden layers are initialized by using the dataset. Only the parameters in the output layer are trained by the conventional linear regression.}
  \label{fig1}
\end{figure}

The SWIM algorithm was proposed in \cite{Bolager2023}, which initializes the parameters of a neural network. Figure~\ref{fig1} shows the framework of the SWIM algorithm. In the SWIM algorithm, directly sampled pairs of input data initialize the parameters of all layers except the output layer. Then, the conventional linear regression between outputs of the neural network and the target data determines the parameters of the output layer.

Of course, it is possible to apply the learning based on the gradient descent method after the initialization by the SWIM algorithm. However, as we will see in Sect.~\ref{sec:experiments}, only the initialization by the SWIM algorithm yields reasonable performance. Hence, the SWIM algorithm reduces the computational cost of the learning process.

Below, we provide a brief explanation of the SWIM algorithm, with the flow outlined in Algorithm~1. Let $X \subset \mathbb{R}^D$ be the input space with the Euclidean norm $\|\cdot\|$ and the inner product $\langle \cdot, \cdot \rangle$. Furthermore, let $\Phi$ be a neural network with $L$ hidden layers; the activation function is $\phi: \mathbb{R} \rightarrow \mathbb{R}$, and the network parameters in the $l$-th layer are $\{W_l, \bm{b}_l\}$ for $l=1$ to $l=L+1$. For $\bm{x} \in X$, we write a result from the activation function with the $i$-th element $[W_l \Phi^{(l-1)}(\bm{x}) - \bm{b}_l]_i$ as $[\Phi^{(l)}(\bm{x})]_i = \phi([W_l \Phi^{(l-1)}(\bm{x}) - \bm{b}_l]_i)$, where $\Phi^{(0)}(\bm{x}) = \bm{x}$. The number of nodes in the $l$-th layer is denoted as $N_l$ with $N_0 = D$; $N_{L+1}$ is the output dimension. $W_l$ has a matrix form, and we write the $i$-th row of $W_l$ as $\bm{w}_{l,i}$; the $i$-th element of the vector $\bm{b}_l$ is denoted as $b_{l,i}$.

Here, assume that we have a sufficiently large dataset. In the SWIM algorithm, an important step is the sampling procedure from a dataset. Let $\big\{\big(\bm{x}^{(1)}_{0,i},\bm{x}^{(2)}_{0,i}\big)\big\}^{N_{l}}_{i=1}$ for $l=1,\ldots ,L$ be the sampled pairs of data points over $X \times X$. That is, to determine the weight and bias of a single node in a layer, we sample a pair consisting of two data points; we avoid using the same data as a pair, i.e., $\bm{x}^{(1)}_{0,i} \neq \bm{x}^{(2)}_{0,i}$. Using the sampled data pair, the weight and bias are initialized as follows:
\begin{align}
  \bm{w}_{l,i}=s_{1}\frac{\widetilde{\bm{x}}^{(2)}_{l-1,i}-\widetilde{\bm{x}}^{(1)}_{l-1,i}}{||\widetilde{\bm{x}}^{(2)}_{l-1,i}-\widetilde{\bm{x}}^{(1)}_{l-1,i}||^{2}}, \quad
  b_{l,i}= \langle \bm{w}_{l,i},\widetilde{\bm{x}}^{(1)}_{l-1,i} \rangle + s_{2},
  \label{eq:parameters}
\end{align}
where $s_1, s_2 \in \mathbb{R}$ are hyperparameters, and $\widetilde{\bm{x}}^{(j)}_{l-1,i}=\Phi^{(l-1)}(\bm{x}^{(j)}_{0,i})$ for $j=1,2$. The output layer is selected by the conventional linear regression, and the parameters $W_{L+1}$ and $\bm{b}_{L+1}$ are set as follows:
\begin{align}
  W_{L+1}, \bm{b}_{L+1} 
= \underset{\widetilde{W}_{L+1}, \widetilde{\bm{b}}_{L+1}} {\operatorname{argmin}} \, \mathcal{L} \left(\widetilde{W}_{L+1} \Phi^{(L)}(X) - \widetilde{\bm{b}}_{L+1}, Y\right),
\end{align}
where $\mathcal{L}$ is the loss function, which is usually set as a mean squared error.

\begin{figure}[!t]
\begin{algorithm}[H]
  \caption{The SWIM algorithm}
  \begin{algorithmic}[1]\label{alg:swim}
    \STATE \textbf{Constants:} $\varepsilon  \in \mathbb{R}_{>0}$, $\varsigma \in \mathbb{N}_{>0}, L \in \mathbb{N}_{>0}, \{N_l \in \mathbb{N}_{>0}\}_{l=1}^{L+1}, s_1, s_2 \in \mathbb{R}$
    \STATE \textbf{Data:} $X = \{\bm{x}_i \in \mathbb{R}^D\}_{i=1}^M$, $Y = \{\bm{f}(\bm{x}_i) = \bm{y}_i \in \mathbb{R}^{N_{L+1}}\}_{i=1}^M$
    \FOR{$l = 1,2, \ldots ,L$}
      \STATE $\widetilde{M}_{l} \leftarrow \varsigma \cdot \left \lceil \frac{N_l}{M} \right\rceil \cdot M$
      \STATE $P^{(l)}_{i} \leftarrow 0$ \, for \, $i = 1, \dots, \widetilde{M}_{l}$.
      \STATE $\widetilde{X} = \left\{ (\bm{x}_{0,i}^{(1)}, \bm{x}_{0,i}^{(2)}) \mid \Phi^{(l-1)}(\bm{x}_{0,i}^{(1)}) \neq \Phi^{(l-1)}(\bm{x}_{0,i}^{(2)}) \right\}_{i=1}^{\widetilde{M}_{l}}$ \\ $\quad  \sim$ Uniform $(X \times X)$ \quad (where $\Phi^{(0)}(\bm{x}) = \bm{x}$)
      \FOR{$i = 1,2, \ldots ,\widetilde{M}_{l}$}
        \STATE $\widetilde{\bm{x}}_{l-1,i}^{(1)} \leftarrow \Phi^{(l-1)}(\bm{x}_{0,i}^{(1)})$, \quad
$\widetilde{\bm{x}}_{l-1,i}^{(2)} \leftarrow \Phi^{(l-1)}(\bm{x}_{0,i}^{(2)})$
        \STATE $\widetilde{\bm{y}}_{i}^{(1)} \leftarrow f(\bm{x}_{0,i}^{(1)})$, \quad
$\widetilde{\bm{y}}_{i}^{(2)} \leftarrow f(\bm{x}_{0,i}^{(2)})$
        \STATE $\displaystyle P_i^{(l)} \leftarrow \frac{\|\widetilde{\bm{y}}_{i}^{(2)} - \widetilde{\bm{y}}_{i}^{(1)}\|_{L^\infty}}{\max\{\|\widetilde{\bm{x}}_{l-1,i}^{(2)} - \widetilde{\bm{x}}_{l-1,i}^{(1)}\|, \varepsilon\}}$ \label{P}
      \ENDFOR
      \STATE $W_l \in \mathbb{R}^{N_l \times N_{l-1}}$, $b_l \in \mathbb{R}^{N_l}$
      \FOR{$i = 1,2 \ldots ,N_l$}
        \STATE Sample $(\bm{x}_{0,i}^{(1)}, \bm{x}_{0,i}^{(2)})$ from $\widetilde{X}$ with replacement and with probability proportional to $\{P_i^{(l)}\}$;
        \STATE $\widetilde{\bm{x}}_{l-1,i}^{(1)} \leftarrow \Phi^{(l-1)}(\bm{x}_{0,i}^{(1)})$, \quad
$\widetilde{\bm{x}}_{l-1,i}^{(2)} \leftarrow \Phi^{(l-1)}(\bm{x}_{0,i}^{(2)})$
        \STATE $\displaystyle \bm{w}_{l,i} \leftarrow s_1 \frac{\widetilde{\bm{x}}_{l-1,i}^{(2)} - \widetilde{\bm{x}}_{l-1,i}^{(1)}}{\|\widetilde{\bm{x}}_{l-1,i}^{(2)} - \widetilde{\bm{x}}_{l-1,i}^{(1)}\|^2}$, $\,\, b_{l,i} \leftarrow \langle \bm{w}_{l,i}, \widetilde{\bm{x}}_{l-1,i}^{(1)} \rangle + s_2$ \label{W,b}
      \ENDFOR
      \STATE $[\Phi^{(l)}(\bm{x}_{k})]_i \leftarrow \phi([W_l \Phi^{(l-1)}(\bm{x}_k) - \bm{b}_l]_i)$ \\
\qquad \qquad \qquad for $k=1,2,\dots,M$ and $i=1,2,\dots,N_l$
    \ENDFOR
    \STATE $\displaystyle W_{L+1}, \bm{b}_{L+1} \leftarrow \underset{\widetilde{W}_{L+1}, \widetilde{\bm{b}}_{L+1}} {\operatorname{argmin}} \, \mathcal{L}(\widetilde{W}_{L+1} \Phi^{(L)}(X)-\widetilde{\bm{b}}_{L+1}, Y)$
    \RETURN $\{W_l, \bm{b}_l\}_{l=1}^{L+1}$
  \end{algorithmic}
\end{algorithm}
\end{figure}

Here, we should focus on the choice of the data pairs. In \cite{Bolager2023}, there is the following notice, i.e., ``putting emphasis on points that are close and differ a lot with respect to the output of the true function works well.'' Then, the following sampling framework at each layer is employed in the SWIM algorithm. First, we prepare enough data pairs for the $l$-th layer, i.e., $\big\{\big(\bm{x}^{(1)}_{0,i},\bm{x}^{(2)}_{0,i}\big) \big\}_{i=1}^{\widetilde{M}_{l}}$; where $\widetilde{M}_{l}$ is the number of the generated data pairs, and $\widetilde{M}_{l}$ is calculated by $\widetilde{M}_{l} = \varsigma \cdot \left \lceil \frac{N_l}{M} \right\rceil \cdot M$ where $M$ is the number of training data points and $\varsigma \in \mathbb{N}_{>0}$ is a hyperparameter. Note that we avoid the choice with $\Phi^{(l-1)}\big(\bm{x}_{0,i}^{(1)}\big) \neq \Phi^{(l-1)}\big(\bm{x}_{0,i}^{(2)}\big)$. Second, we assign the following quantity to each data pair:
\begin{align}
 P_i^{(l)} = 
\frac{\|\widetilde{\bm{y}}_{i}^{(2)} - \widetilde{\bm{y}}_{i}^{(1)}\|_{L^\infty}}{\max \left\{\|\widetilde{\bm{x}}_{l-1,i}^{(2)} - \widetilde{\bm{x}}_{l-1,i}^{(1)}\|, \varepsilon\right\}},
\label{eq_P}
\end{align}
where $\widetilde{\bm{x}}_{l-1,i}^{(j)} = \Phi^{(l-1)}\big(\bm{x}_{0,i}^{(j)}\big)$ and $\widetilde{\bm{y}}_{i}^{(j)} = f\big(\bm{x}_{0,i}^{(j)}\big)$ for $j = 1,2$, with $f$ denoting the true target function. Here, $\varepsilon > 0$ for $l \in \{1,2,\ldots,L\}$ and $\varepsilon = 0$ for $l = 1$. Note that the $L^{\infty}$ norm is used for the numerator in Eq.~\eqref{eq_P}, which returns the maximum absolute value of the element in the vector $\widetilde{\bm{y}}_{i}^{(2)} - \widetilde{\bm{y}}_{i}^{(1)}$. The parameter $\varepsilon$ is introduced to avoid division by zero. Then, the weights and biases are determined via Eq.~\eqref{eq:parameters} by using a data pair sampled with probability proportional to $\{P_i^{(l)}\}$.

\subsection{Hierarchical features in a neural network for a toy problem}

CNNs are widely used in image recognition tasks and are known to capture low-frequency components in the early-stage hidden layers and high-frequency components in the late-stage hidden layers, as written in Sect.~\ref{sec:introduction}. We refer to the property as hierarchical features.

\begin{table}[t]
\caption{Setting in the preliminary experiment to see the hierarchical features.}
\label{table_1}
\begin{tabular}{lc}
\hline\hline
Input dimension &  1\\
\# of hidden layers & 4 \\
\# of nodes in each hidden layer & 128\\
Output dimension & 1\\
\# of epochs & 1000\\
Activation function & $\sin$\\
Optimization method & Adam\\
Loss function & Mean Squared Error\\
Learning rate & $0.01$\\
\hline\hline
\end{tabular}
\end{table}

Here, we investigate whether fully connected neural networks also exhibit the hierarchical features. Note that the fully connected neural networks used here do not include any convolution layers. As a toy problem, we perform learning of a simple one-dimensional function $f(x) = \sin(4 \pi x)$ by fully connected neural networks with four hidden layers. The experimental setting is shown in Table~\ref{table_1}.

\begin{figure}[tb]
  \centering
  \includegraphics[width=0.48\textwidth]{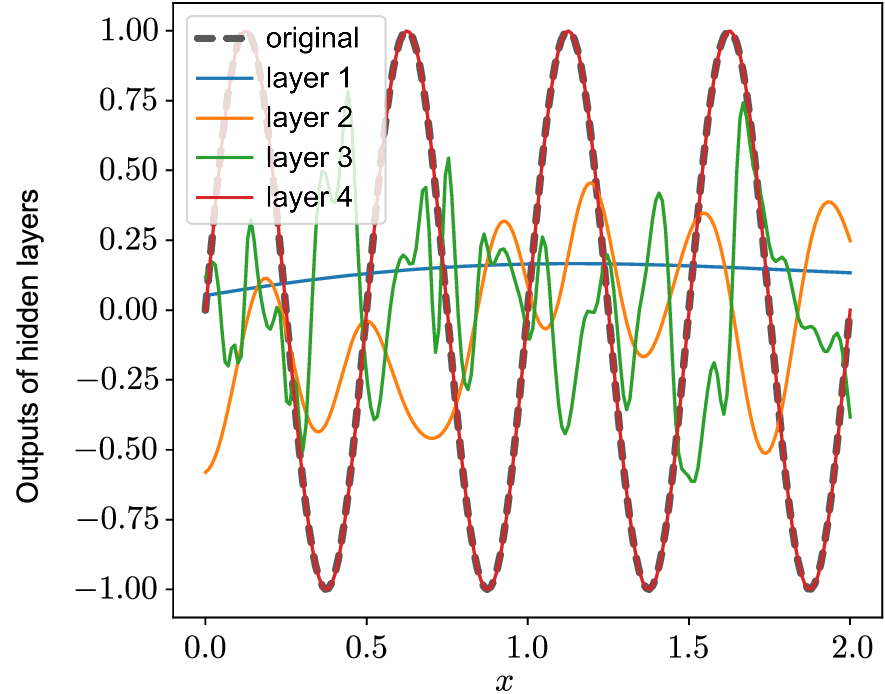}
  \caption{(Color online) Outputs of the trained fully connected neural network for each layer. The horizontal axis is the input $x$, and the vertical axis represents values with a matrix product of the weights of the final layer $W_{L+1}$ to each hidden layer. The dashed curve corresponds to the original function $f(x)$. Since the learning effectively finished, the output of the final layer, i.e., ``layer 4'', matches the original function $f(x)$ well.
}
  \label{fig:FCNN}
\end{figure}

To investigate the role of each hidden layer, we directly connect the final weight matrix $W_{L+1}$ to each hidden layer and visualize its output like using a deconvnet in CNNs. Thus, we define the output of each layer as
\begin{align}
  \mathrm{layer}~l : \,\, W_{L+1} \Phi^{(l)}(\bm{x}) - \bm{b}_{L+1}, \quad l=1,2,3,4.
\end{align}
Figure~\ref{fig:FCNN} depicts the results. The outputs of layer 4, i.e., the final outputs of the neural network, closely match the original function $f(x)$, confirming that the learning was successful. We can see that the curves gradually become more complex in the order of layers 1, 2, and 3. Hence, the early-stage hidden layers of the network learn the feature of low-frequency components; this fact indicates that the fully connected neural network exhibits the property of the hierarchical features, similar to CNNs.

\section{Proposal to employ information on hierarchical features}
\label{sec:proposal}

\subsection{Revisit on the role of hyperparameters $s_1$}

As denoted in Sect.~\ref{sec:background_swim}, the original SWIM algorithm has two hyperparameters, $s_1$ and $s_2$, used to determine the weights and biases, respectively; see Eq.~\eqref{eq:parameters}. In \cite{Bolager2023}, these values are determined with the following discussions. 

The weight parameter $\bm{w}_{l,i}$ is determined by the data pair, as in Eq.~\eqref{eq:parameters}. Intuitively, the weight vector is in the direction of the difference between the two points, and the inner product from a reference point in that direction yields the bias. Consequently, the selection of the parameters described above enables the effective utilization of nonlinear transformations. To see this construction, for simplicity, we here take 
\begin{align}
\widetilde{\bm{x}}^{(2)}_{l-1,i} = -\widetilde{\bm{x}}^{(1)}_{l-1,i}.
\end{align}
Then, we have
\begin{align}
\bm{w}_{l,i}
= 
- \frac{s_{1}}{2}\frac{ \widetilde{\bm{x}}^{(1)}_{l-1,i}}{||\widetilde{\bm{x}}^{(1)}_{l-1,i}||^{2}}
\end{align}
and
\begin{align}
b_{l,i}
= -\frac{s_{1}}{2} \frac{1}{||\widetilde{\bm{x}}^{(1)}_{l-1,i}||^{2}}
\langle \widetilde{\bm{x}}^{(1)}_{l-1,i}, \widetilde{\bm{x}}^{(1)}_{l-1,i} \rangle + s_{2} 
= -\frac{s_{1}}{2} + s_{2}.
\end{align}
Hence, the value of the activation function on $\widetilde{\bm{x}}^{(1)}_{l-1,i}$ becomes
\begin{align}
\phi\left( \langle \bm{w}_{l,i} , \widetilde{\bm{x}}^{(1)}_{l-1,i} \rangle -b_{l,i} \right) 
&= 
\phi\left( \left\langle - \frac{s_{1}}{2}\frac{ \widetilde{\bm{x}}^{(1)}_{l-1,i}}{||\widetilde{\bm{x}}^{(1)}_{l-1,i}||^{2}} , \widetilde{\bm{x}}^{(1)}_{l-1,i} \right\rangle
 -\frac{s_{1}}{2} + s_{2}
\right) \nonumber \\
&=
\phi\left( - s_{1}+ s_{2} \right).
\label{eq_role_s1}
\end{align}
Note that the activation function takes $\phi(s_{2})$ on $\widetilde{\bm{x}}^{(2)}_{l-1,i}$. For example, we consider $\tanh$ as the activation function. Then, if we set $s_{1} = \ln 3$ and $s_2 = (\ln 3) / 2$, the activation function takes $-0.5$ and $+0.5$ on $\widetilde{\bm{x}}^{(1)}_{l-1,i}$ and $\widetilde{\bm{x}}^{(2)}_{l-1,i}$, respectively. Hence, these hyperparameters effectively utilize the nonlinear region of $\tanh$. The similar discussions were used to determine the values of the hyperparameters in \cite{Bolager2023}. 

Based on the above discussion, we conjecture that $s_1$ is more relevant than $s_2$ because $s_2$ is related only to the bias term. Hence, we performed several preliminary numerical experiments to investigate the effects of these hyperparameters; the numerical results suggested that $s_1$ significantly affects the final performance. By contrast, $s_2$ has little effect on performance. Hence, we focus primarily on the hyperparameter $s_1$ in the following discussion.

\begin{figure}[tb]
  \begin{center}
  \includegraphics[width=0.45\textwidth]{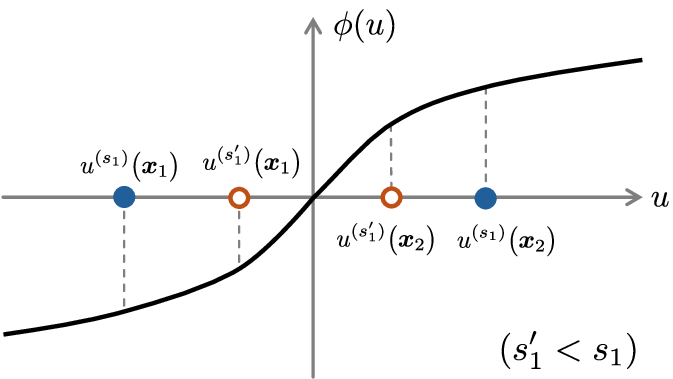}
  \caption{(Color online) Example of different choices with the parameter for $s_{1}$. Even if the same data pair $(\bm{x}_{1}, \bm{x}_{2})$ is used, the degree of nonlinearity utilized varies depending on the parameters. When $s'_1 < s_{1}$, the small parameter $s'_1$ utilizes only the portion of the activation function.}
  \label{fig:nonlinear}
  \end{center}
\end{figure}

To discuss the role of $s_1$, we define the following quantity:
\begin{align}
u^{(s_1)}(\bm{x}) =
\left\langle s_1 \frac{\widetilde{\bm{x}}^{(2)}_{l-1,i}-\widetilde{\bm{x}}^{(1)}_{l-1,i}}{||\widetilde{\bm{x}}^{(2)}_{l-1,i}-\widetilde{\bm{x}}^{(1)}_{l-1,i}||^{2}} , \bm{x} \right\rangle,
\end{align}
which is one of the arguments in the activation function, i.e., $\phi(u^{(s_1)}(\bm{x}) + b)$. As a pedagogical example, we will consider the role of parameters here using $\tanh$ as the activation function. Figure~\ref{fig:nonlinear} shows the shape of the activation function. Since the parameter $s_1$ varies the weight parameter vector $\bm{w}_{l,i}$ in Eq.~\eqref{eq_role_s1}, the actual coordinate $u^{(s_1)}(\bm{x})$ depends on the value of $s_1$ even for the same input vector $\bm{x}$. In Fig.~\ref{fig:nonlinear}, the same data points, $\bm{x}_1$ and $\bm{x}_2$, yield different coordinates, $u^{(s_i)}(\bm{x}_1)$ and $u^{(s_i)}(\bm{x}_2)$, for the different parameters $s_1$ and $s_1'$.

From the above discussion, we obtain the following insights. A smaller parameter $s_1$ tends to focus on narrow regions of the activation function, where its behavior is approximately linear. As a result, one utilizes only these nearly linear regions. Since linear representations are well-suited for approximating low-frequency components, this setting promotes the learning of low-frequency information.

By contrast, a larger parameter $s_1$ will map the input to a larger region of the activation function's output space, thereby making more use of the nonlinearity part of the activation function. Hence, the network can capture subtle differences among input values. Such behavior is suitable for learning high-frequency information.

From these observations, we conclude that the hyperparameter $s_1$ controls the degree of nonlinearity exploited by the activation function and is therefore closely related to the frequency components of the information captured by the network.

\subsection{Proposal of new parameter initialization method}

\begin{figure}[tb]
  \begin{center}
  \includegraphics[width=0.48\textwidth]{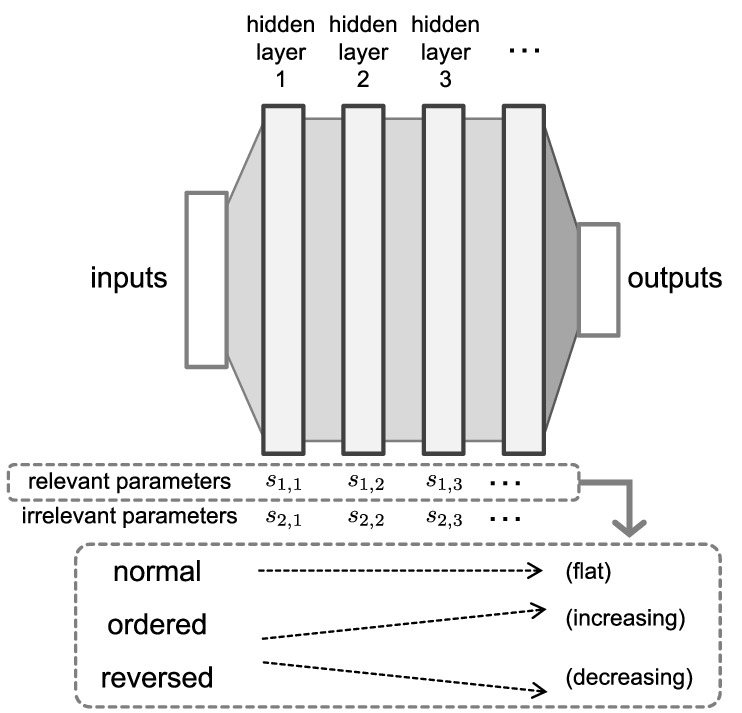}
  \caption{Framework to include the information on the hierarchical features. The relevant hyperparameter $s_1$ can vary across hidden layers. The ``ordered'' is the proposed method, in which $s_{1,l}$ gradually increases with the layer index $l$. By contrast, $s_{1,l}$ gradually decreases in the ``reversed'' method. The ``normal'' corresponds to the original SWIM algorithm in which $s_{1,l}$ is constant.
}
  \label{fig:proposal}
  \end{center}
\end{figure}

According to the above discussions, we propose a framework to include the information on the hierarchical features. In the original SWIM algorithm, the scale factor $s_1$ is set to a constant value for all hidden layers. However, since $s_1$ controls the scale of the activation function, we change $s_1$ for each intermediate layer, as in Fig.~\ref{fig:proposal}; we write the hyperparameter for the $l$-th layer as $s_{1,l}$.

The discussion on the hierarchical features suggests that it would be better to process coarse information in the early-stage hidden layers and deal with fine details in the late-stage hidden layers. Hence, we propose the following ``ordered'' setting:
\begin{align}
& \bm{s}_1 = [s_{1,1}, s_{1,2},\cdots , s_{1,L}], \nonumber \\
& \bm{s}_2 = [s_{2,1}, s_{2,2}, \cdots, s_{2,L}] = \frac{1}{2}\bm{s}_1,
\end{align}
where $s_{1,1} \leq s_{1,2} \leq s_{1,3} \leq \cdots \leq s_{1,L}$. While we tried several settings for $s_2$, there were no significant differences. Hence, we employed the above setting.

In Fig.~\ref{fig:proposal}, we depict two other methods; the ``reversed'' method has the reversed order of $s_{1,l}$, which is designed to capture high-frequency components in the early-stage hidden layers and gradually decrease the scale factor in the late-stage hidden layers. The ``normal'' is the original SWIM algorithm; $s_{1,l}$ is constant for all the indices $l$.

\section{Numerical Experiments}
\label{sec:experiments}

In this section, we demonstrate the effects of the information on the hierarchical features. Here, we compare the proposed method with the original SWIM algorithm and the reversed method. 

Although we checked various settings in preliminary numerical experiments, we set the parameters to $\varepsilon = 0.01$, $\varsigma = 10$, and the activation function to $\phi(\cdot) = \sin(\cdot)$ in all the numerical results in this paper. In addition, for the original SWIM algorithm, we show numerical results with $s_{1,i} = \ln{3}, s_{2,i} = (\ln{3})/2$ for all index $i$.

\subsection{Regression task on one-dimensional data}
\label{sec:experiments_1D}
We first consider a regression task on one-dimensional data. We use the following target function:
\begin{align}
  f(x) =& \sin(2 \cdot 2 \pi x) +  0.3 \sin(20 \cdot 2 \pi x) \nonumber \\
       & + 0.1 \sin(30 \cdot 2 \pi x) + 0.05 \sin(40 \cdot 2 \pi x).
\label{eq_1D_true}
\end{align}
We sample 200 data points uniformly from the interval $[0, 2]$ and use these points as the input data. The target values are computed using the above function. Table~\ref{table_2} shows the experimental setting. The number of hidden layers is $4$, and all the hidden layers have the same number of nodes; we change the number of nodes in the experiments. The hyperparameter $\bm{s}_1$ is set as $\{s_{1,1}, s_{1,2}, s_{1,3}, s_{1,4}\} = \{0.5, 7, 13.5, 20\}$; only the increasing tendency is crucial, and there is no reason to select these hyperparameter values. We tried several hyperparameters, and all the results have the same tendency. Since the above hyperparameter value appears to yield reasonable results, we present the corresponding numerical results for the above hyperparameter values. In addition, note that we do not apply the backpropagation procedure; only the samplings and the conventional linear regression on the final layer are employed.

\begin{table}[b]
\caption{Experimental setting for the one-dimensional problem.}
\label{table_2}
\begin{tabular}{lc}
\hline\hline
Input dimension &  1\\
\# of hidden layers & 4 \\
\# of nodes in each hidden layer \qquad\qquad & (variable)\\
Output dimension & 1\\
\hline\hline
\end{tabular}
\end{table}

\begin{figure}[t]
  \begin{center}
  \includegraphics[width=0.48\textwidth]{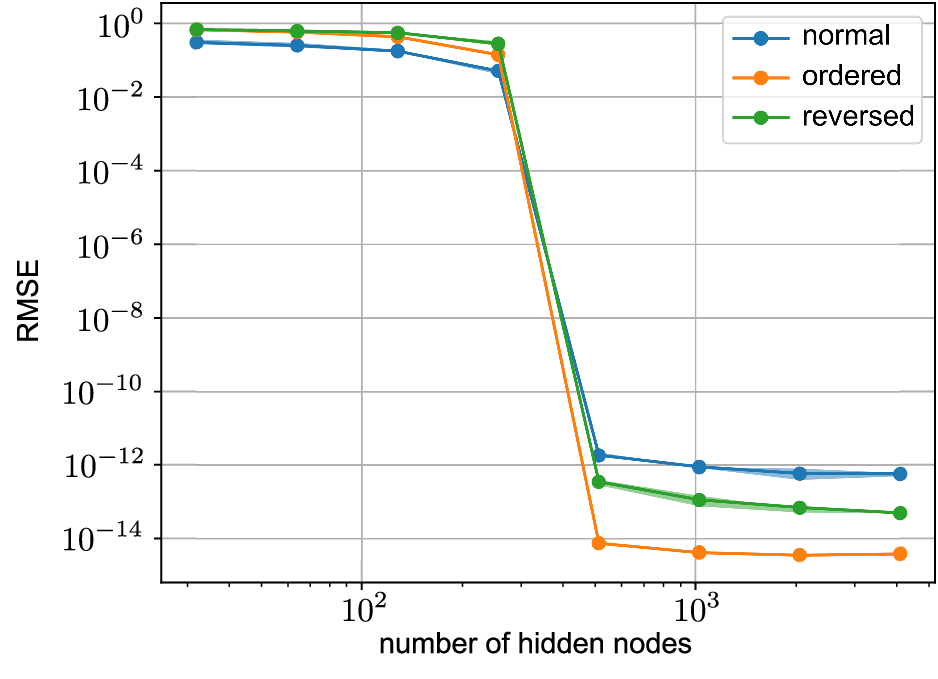}
  \caption{(Color online) 
The RMSE and standard deviation of the neural network outputs for each method, averaged over five runs. The horizontal axis represents the number of nodes in each hidden layer, and the vertical axis corresponds to the RMSE. Error bars based on standard deviation are present but are too small to be discernible.}
  \label{fig:ex_1d_rmse}
  \end{center}
\end{figure}

Figure~\ref{fig:ex_1d_rmse} depicts the root mean square error (RMSE) of the output of the neural network for each method. In the ``normal'' method, the original SWIM algorithm is used, and the hyperparameters are common across all layers. In the ``reversed'' method, the hyperparameter $\bm{s}_1$ has the reversed order, i.e., $\{s_{1,1}, s_{1,2}, s_{1,3}, s_{1,4}\} = \{20, 13.5, 7, 0.5\}$; each element of the vector $\bm{s}_2$ is half the value of that of $\bm{s}_1$.

When the number of nodes in the hidden layers is small, there are insufficient nodes for learning, which results in poor accuracy across all methods. When there is a sufficient number of nodes, Fig.~\ref{fig:ex_1d_rmse} indicates that the proposed method (ordered) achieves the lowest RMSE values compared to the ``normal'' method (the original SWIM algorithm) and the ``reversed'' method.

\begin{figure*}[tb]
  \centering
  \includegraphics[width=0.9\textwidth]{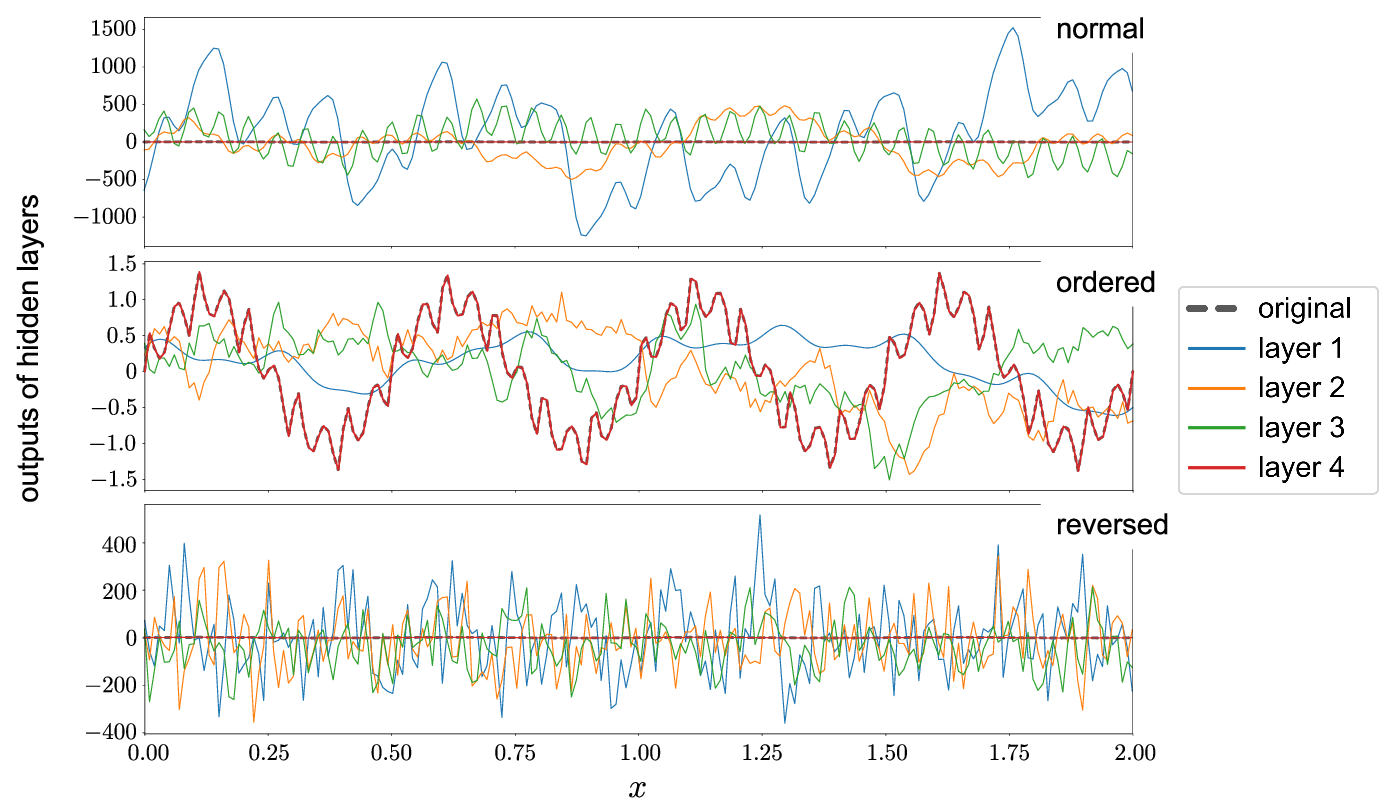}
  \caption{(Color online) The output values computed by connecting the final layer directly to each hidden layer. The dashed lines correspond to the true function $f(x)$ in Eq.~\eqref{eq_1D_true}. The values of layer 4 are the outputs of the neural network, and hence, they mostly match the true function $f(x)$. Note that due to the large scale of the vertical axes in the top (normal) and bottom (reversed) figures, the true function $f(x)$ appears almost constant in these figures.
}
  \label{fig:ex_1d_out}
\end{figure*}

To discuss the reason for the good performance of the proposed method, we next see the role of each hidden layer, as in Fig.~\ref{fig:FCNN}. Figure~\ref{fig:ex_1d_out} depicts the output of the neural network for each method; connecting the final layer directly to each hidden layer yields the output values. Here, we show the results for the neural network with 1024 nodes in all hidden layers. One of the remarkable features in Fig.~\ref{fig:ex_1d_out} is that the value of the vertical axis of the ordered method is close to the target function. However, the values of the vertical axis of the normal and the reversed methods are not close to the target function. Furthermore, the ordered method captures the low-frequency components in the early-stage hidden layers, which would allow the network to learn effectively. In Appendix~\ref{sec:appendix}, we show the histograms of the $\lvert \bm{x}_2-\bm{x}_1 \rvert$ and confirm that the proposed method (``ordered'') is effective for capturing the hierarchical features.

\subsection{Classification task on MNIST}

Next, we apply the proposed method to a classification task on the MNIST dataset \cite{LeCun1998,Deng2012}. The number of training images of handwritten digits is 15,000, and that of test images is 10,000. Table~\ref{table_3} shows the experimental setting. The hyperparameter $\bm{s}_1$ is set as $\{s_{1,1}, s_{1,2}, s_{1,3}, s_{1,4}\} = \{0.01, 0.173, 0.337, 0.5\}$. Again, note that we do not apply the backpropagation procedure; only the samplings and the conventional linear regression on the final layer are employed.

\begin{table}[b]
\caption{Experimental setting for the MNIST problem.}
\label{table_3}
\begin{tabular}{lc}
\hline\hline
Input dimension &  784 (28$\times$28 pixels)\\
\# of hidden layers & 4 \\
\# of nodes in each hidden layer \qquad & (variable)\\
Output dimension & 10 (10 digits)\\
\hline\hline
\end{tabular}
\end{table}

\begin{figure}[tb]
  \centering
  \includegraphics[width=0.48\textwidth]{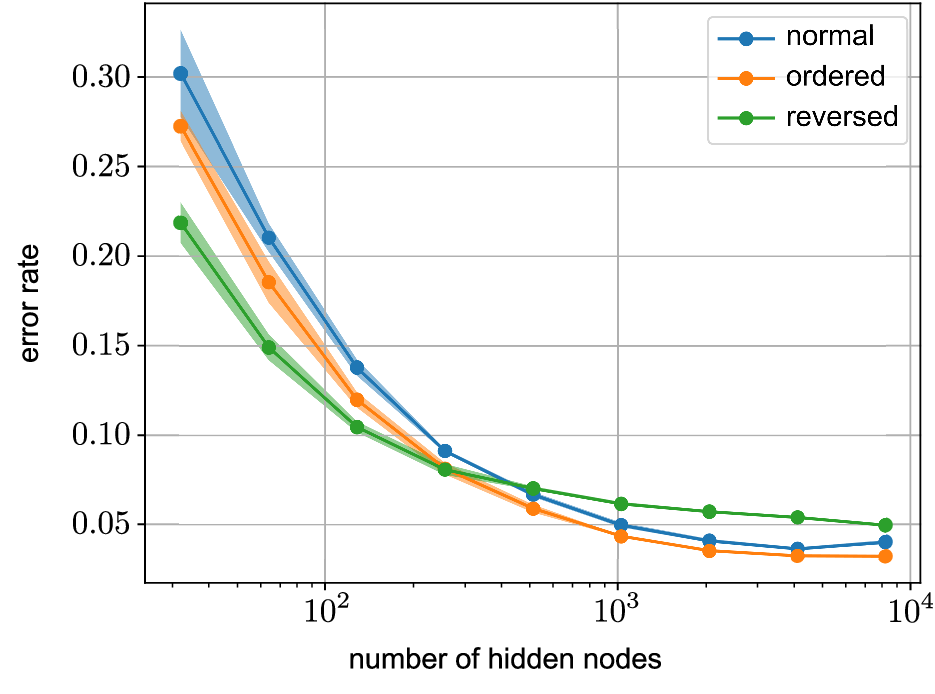}
  \caption{(Color online) The error rate and standard deviation of the neural network outputs for each method, averaged over five runs. The horizontal axis is the number of nodes in each hidden layer, and the vertical axis corresponds to the error rate.}
  \label{fig:ex_mnist_rmse}
\end{figure}

Figure~\ref{fig:ex_mnist_rmse} depicts the error rate of the neural network outputs for each method. In the classification task, the proposed method (ordered) achieves the lowest error rates in cases of large numbers of nodes. Hence, even in the high-dimensional inputs, the information on the hierarchical features improves the performance.

Here, we comment on why the proposed method (ordered) performs worse than the reversed method when the number of nodes in the hidden layers is small. As we discussed in Sec.~III.~A, a small value of the hyperparameter $s_1$ mainly focuses on coarse information (low-frequency). Since a small number of nodes in the hidden layers cannot deal with the coarse information sufficiently, we observe poor performance. By contrast, a larger $s_1$ allows us to deal with detailed information. Note that the small number of nodes in the hidden layers corresponds to rough information processing. Then, it would be natural that the rough information processing of ``detailed information'' yields better performance than the rough information processing of ``coarse information.''

Of course, the above discussion is valid only for the small number of nodes in the hidden layers, where the error rate is inherently high and performance is poor. As far as we checked with several different examples, activation functions and hyperparameter settings, the information on the hierarchical features works well when there is a large number of nodes in the hidden layer.

\section{Conclusion}
\label{sec:conclusion}

The data-driven initialization of weights and biases enables us to avoid learning with conventional backpropagation. In this work, we proposed an initialization method to leverage information on the hierarchical features. The crucial point is the role of the hyperparameter $s_1$ in the original SWIM algorithm, which changes the role of nonlinearity in the activation function. The gradual change of the hyperparameter enables us to capture the low-frequency components in the early-stage hidden layers. We demonstrated the effects of information on the hierarchical features by numerical experiments, and the proposed method outperforms the original SWIM method and the reversed-order method in both regression and classification tasks.

There are several remaining tasks in the future. First, applying this approach to other types of neural networks, such as CNNs and recurrent neural networks (RNNs), is an important future task. Second, further investigation of how to set the scaling factor $s_1$ is also important. As far as we checked, the gradual change of parameters improves performance. However, the degree of improvement varies depending on the settings. Optimal parameter settings may exist. Furthermore, while this study varied parameters according to layer depth, practical adjustments could be possible; for example, the parameter change within the same hidden layer might be effective in some situations.

Studies utilizing features such as the hierarchical features in sampling-based initialization techniques have just started. We hope this study will contribute to the exploration of efficient learning methods for future neural networks.

\begin{acknowledgments}

This work was supported by JSPS KAKENHI Grant Number JP21K12045.
\end{acknowledgments}

\appendix

\section{Histogram of $|\bm{x}_{l,i}^{(2)} - \bm{x}_{l,i}^{(1)}|$} 
\label{sec:appendix}

\begin{figure*}[tb]
  \centering
  \includegraphics[width=0.9\textwidth]{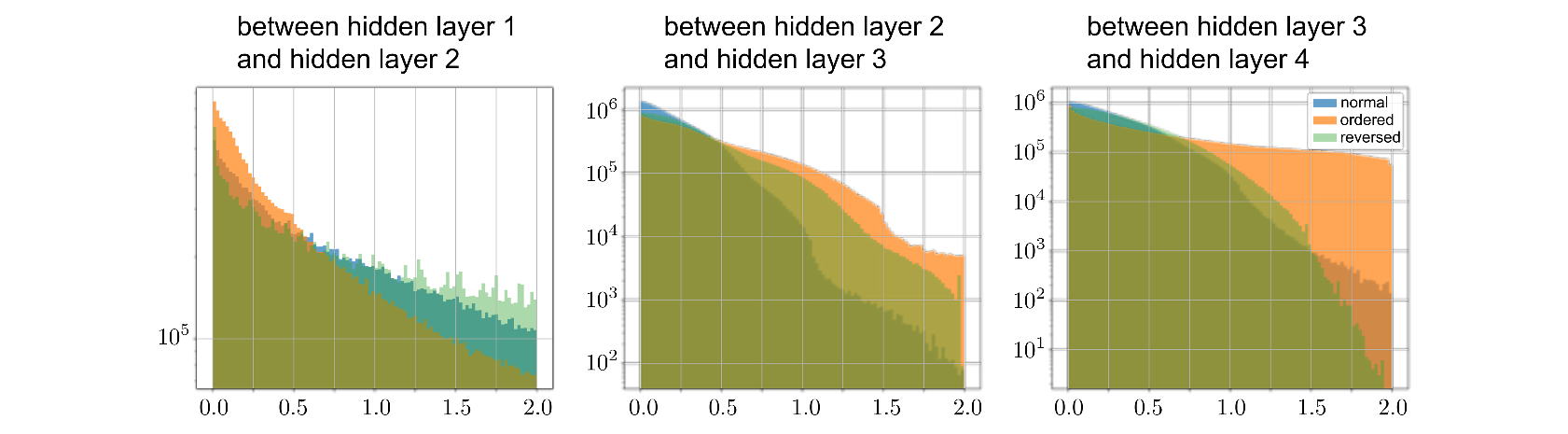}
  \caption{(Color online) The histogram of $|\bm{x}_{l,i}^{(2)} - \bm{x}_{l,i}^{(1)}|$ for each method in each hidden layer pair. The left, middle, and right figures correspond to the histograms for the parameter samplings between the hidden layers 1 and 2, 2 and 3, and 3 and 4, respectively.
  }
  \label{fig:dist_x2x1}
\end{figure*}

In this appendix, we show the histogram of $|\bm{x}_{l,i}^{(2)} - \bm{x}_{l,i}^{(1)}|$ after sampling for each method in the one-dimensional regression task in Sect.~\ref{sec:experiments_1D}. The settings are similar to those in the main text, but the number of hidden nodes is set to 1024. Figure~\ref{fig:dist_x2x1} depicts the histograms of $|\bm{x}_{l,i}^{(2)} - \bm{x}_{l,i}^{(1)}|$ for each hidden layer. Note that there are $1024 \times 1024$ edges between each pair of hidden layers, and we constructed $20$ different realizations of neural networks. Hence, we performed totally $1024 \times 1024 \times 20 = 20,971,520$ samplings of $\bm{x}_{l,i}^{(1)}$ and $\bm{x}_{l,i}^{(2)}$ for each pair of hidden layers. The histograms in Fig.~\ref{fig:dist_x2x1} were constructed from these sampling results.

Note that a small value of $|\bm{x}_{l,i}^{(2)} - \bm{x}_{l,i}^{(1)}|$ indicates that the data pair is close to each other in the input space and operates in a nearly linear region. Such data pairs are therefore suitable for capturing low-frequency components. By contrast, large values of $|\bm{x}_{l,i}^{(2)} - \bm{x}_{l,i}^{(1)}|$ correspond to data pairs far from each other, which is appropriate for capturing high-frequency components.

In the first hidden layer, the histogram of ``ordered'' tends to have small values of $|\bm{x}_{l,i}^{(2)} - \bm{x}_{l,i}^{(1)}|$ compared to the other two methods. By contrast, the histogram of ``reversed'' shows a tendency toward larger distances. As the layer depth increases, the histogram of ``ordered'' gradually shifts to have large values of $|\bm{x}_{l,i}^{(1)} - \bm{x}_{l,i}^{(2)}|$ compared to those of the other two methods.

From these results, we confirm that the proposed method effectively utilizes data pairs suitable for capturing low-frequency components in the early-stage hidden layers and high-frequency components in the late-stage hidden layers. Hence, the proposed method effectively captures the hierarchical features.

\end{document}